\documentclass[journal, twoside]{IEEEtran}
\usepackage{graphicx}
\usepackage{caption}
\usepackage{subcaption}
\usepackage{adjustbox}
\usepackage[utf8]{inputenc}
\usepackage{amsmath}
\usepackage[numbers]{natbib}
\usepackage{mathtools}
\ifCLASSINFOpdf
\else
\fi
\begin{document}
%
% paper title
% Titles are generally capitalized except for words such as a, an, and, as,
% at, but, by, for, in, nor, of, on, or, the, to and up, which are usually
% not capitalized unless they are the first or last word of the title.
% Linebreaks \\ can be used within to get better formatting as desired.
% Do not put math or special symbols in the title.
\title{Incident Detection on Junctions Using \\ Image Processing}
%
%
% author names and IEEE memberships
% note positions of commas and nonbreaking spaces ( ~ ) LaTeX will not break
% a structure at a ~ so this keeps an author's name from being broken across
% two lines.
% use \thanks{} to gain access to the first footnote area
% a separate \thanks must be used for each paragraph as LaTeX2e's \thanks
% was not built to handle multiple paragraphs
%

\author{Murat~Tulgaç, Enes~Yüncü, Mohamad-Alhaddad and~Ceylan~Yozgatlıgil% <-this % stops a space
%\thanks{Manuscript received ...}
\thanks{M.Tulgaç, E. Yüncü, and M.Alhaddad are with the ISSD, 06800 Ankara, Turkey (e-mail: murat.tulgac@issd.com.tr; enes.yuncu@issd.com.tr; muhammed@issd.com.tr).}% <-this % stops a space
\thanks{C.Yozgatligil is with the Department of  Statistics  at  METU, 06800 Ankara, Turkey (e-mail:
ceylan@metu.edu.tr).}}% <-this % stops a space
\maketitle

% As a general rule, do not put math, special symbols or citations
% in the abstract or keywords.
\begin{abstract}
In traffic management, it is a very important issue to shorten the response time by detecting the incidents (accident, vehicle breakdown, an object falling on the road, etc.) and informing the corresponding personnel. In this study, an anomaly detection framework for road junctions is proposed. The final judgment is based on the trajectories followed by the vehicles. Trajectory information is provided by vehicle detection and tracking algorithms on visual data streamed from a fisheye camera. Deep learning algorithms are used for vehicle detection, and Kalman Filter is used for tracking. To observe the trajectories more accurately, the detected vehicle coordinates are transferred to the bird's-eye view coordinates using the lens distortion model prediction algorithm. The system determines whether there is an abnormality in trajectories by comAPring historical trajectory data and instantaneous incoming data. The proposed system has achieved \%84.6 success in vehicle detection and \%96.8 success in abnormality detection on synthetic data. The system also works with a \%97.3  success rate in detecting abnormalities on real data. 
\end{abstract}

% Note that keywords are not normally used for peerreview papers.

% For peer review papers, you can put extra information on the cover
% page as needed:
% \ifCLASSOPTIONpeerreview
% \begin{center} \bfseries EDICS Category: 3-BBND \end{center}
% \fi
%
% For peerreview papers, this IEEEtran command inserts a page break and
% creates the second title. It will be ignored for other modes.
\IEEEpeerreviewmaketitle

\section{Introduction}

\IEEEPARstart{T}{he} demand for transportation rises year by year due to the increase in the population. As a result, traffic congestion occurs by exceeding the capacity of existing roads. Traffic congestions are examined in two groups as recurrent and non-recurrent ones. Recurrent congestions are based on a certain routine, and not affected by temporary events. Non-recurrent congestions are experienced due to weather or temporary events in traffic. 

It is critical to inform the operators by identifying accidents, vehicle malfunctions, falling objects, or similar temporary events as soon as possible, especially at intersections. Fortunately, the study of smart monitoring systems has been a great interest in research and development. Remarkable progress has been achieved to deploy traffic monitoring systems that will detect abnormalities at intersections and notify who is in charge of taking action.

In this paper, we address the use of an omnidirectional (fisheye) camera for intersection monitoring. We provide a framework for incident detection which consists of three steps:

\begin{enumerate}
    \item Detect vehicles in the camera Field of View (FoV)
    \item Track the vehicles entering/exiting the FoV
    \item ComAPre their followed paths trajectories with the normal path and detect any abnormality.
\end{enumerate}

In order to detect an anomaly, normal behavior of the specific route must be characterized \cite{Reference1}. Afterward, by monitoring the system behavior continuously, the anomaly can be detected by defining time intervals that do not comply with normal behavior. In the proposed method, the difference between the normal behavior and the behavior of the entering vehicles is evaluated on the trajectories followed by the vehicles. For trajectory extraction, we utilized deep learning techniques to develop our vehicle detector, then we mapped the scene to the bird's-eye view. Finally, we tracked the vehicles using Munkres and Kalman filter methods.

In this study, we propose a framework that detects trajectories on the images taken from a fisheye camera installed at the intersection and detects abnormalities by comAPring these trajectories with the past trajectories. This framework acts as a subsystem for Intelligent Traffic System (ITS). The flow chart of the proposed method is shown below in Figure 1. 

\begin{figure}[!h]
\centering
\includegraphics[height=7.5cm,keepaspectratio]{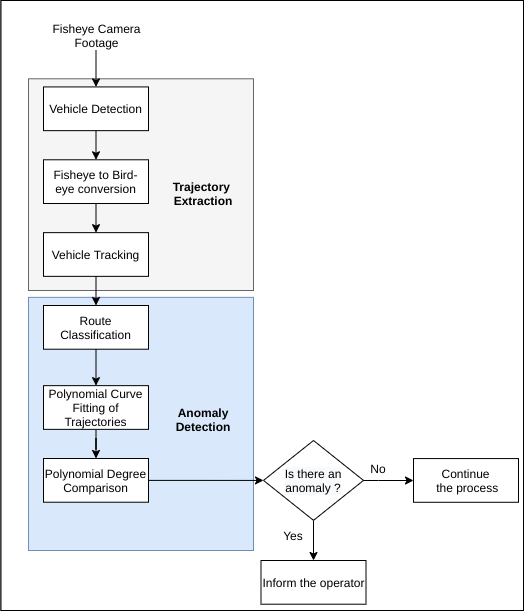}
% where an .eps filename suffix will be assumed under latex, 
% and a .pdf suffix will be assumed for pdflatex; or what has been declared
% via \DeclareGraphicsExtensions.
\caption{Flow chart of the proposed system.}
\end{figure}

The rest of the study is evaluated under four headings: Section-2 addresses the literature advancement in incident detection at intersections. Section-3 illustrates the synthetic and real data used in the study. In the Section-4, the proposed method is explained and sub-steps related to the method are presented. In Section-5, the results of the study on synthetic and real data, conclusions, and summary of the study are presented.

\section{Related Studies}

The omnidirectional cameras are known for their wider Field of View (FoV) than conventional cameras. Thus, using them in monitoring tasks is beneficial in terms of setup and maintenance costs. The studies addressing omnidirectional cameras have reported several challenges faced in capturing a large image and analyze the highly distorted objects at variable scales in real-time.

In their work, Lee et al. \cite{Reference2} have created a vehicle trip table estimation which tracks the origin-destination points of vehicles passing through the intersection. Their tracker relies on particle filters and motion dynamics and has suffered from appearance changes and different weather conditions which makes it limited in robustness. Wang et al. \cite{Reference3} have addressed the issues faced by relying on a fisheye camera for vehicle counting at an intersection, due to severe distortion and perspective effects. They came up with a method based on motion similarity and probabilistic weights to transfer the motion knowledge between long and short-term trajectories between points. Other attempts to surmount the distortion was to track motion based on background subtraction models \cite{Reference4,Reference5,Reference6}.

Other studies have addressed the same problem with standard cameras (monocular vision), and remarkable outputs were reported so far in the literature. We found out that the researchers usually attempt to do background subtraction or feature extraction and tracking variants at the first step for detection purposes, then tracking the vehicle path with other perimeters \cite{Reference7,Reference8,Reference9,Reference10,Reference11}. 

To track the behavior, some studies attempt to rely on the vehicle size ratio in correlation with its direction, velocity, and acceleration obtained from the detection step \cite{Reference12} to understand the onsite situation, while others rely on kernel-based filters to determine whether the distribution of the detected pixels represents an accident or not \cite{Reference13}.  

The main difference between existing attempts and our approach is the lack of incident detection exclusiveness. The current studies focus on the situation where two vehicles may collide, while our approach can detect abnormality based on the passing vehicles' trajectories characteristics. Another aspect is that we embed deep learning trends for ITS applications where addressed research has relied heavily on background subtraction which suffers in terms of robustness, especially for omnidirectional vision.

\section{Data}

Two different data are used in the development and testing phases of the algorithm. These are simulation and real data of vehicles passing the intersection. An algorithm has been developed in which accurate results can be obtained for anomaly detection in both real and simulated data. 

\subsection{Simulation Data}

Simulation data is obtained from the simulation program called VisSim \cite{Reference14}. This data is used in the creation of deviation detection algorithms in trajectory. ComAPred to the fisheye dataset, simulation data has a lower noise rate (Figures 9-10). This caused the trajectories of the vehicles in the simulation data to be much smoother than the actual data. Because of that, new tests were carried out on the fisheye dataset after implementing and testing the algorithm on simulation data to develop a more robust process.

Simulation data is created by simulating the 1-hour flow of a junction of a city in Turkey that is used while the fisheye camera dataset is created. Seven different scenarios (incident at different specific routes, no incidents) are created to test how the system will respond to different situations. In addition to these different scenarios, to characterize the normal behavior, the simulation scenario with no incident was run 30 times. Hence, 37 hours of simulation data are used. In this data, as with tracking algorithms, vehicle-based coordinate data are obtained, and consistency with real data is achieved. The simulation produces several vehicles $V_1, V_2...V_n$ passing through the junction with different followed paths. Each followed path is sampled over several subsequent locations $L_{x,y}$ from the junction entrance until exit. ($x,y$) are the Cartesian coordinates of the vehicle's $V$ location $L$.

After the vehicle-based coordinate data is normalized on all data, the trajectories of the vehicles are sorted by time, and the differences from previous trajectories are observed. Related images for the intersection used in the study and simulation data are shown in Figure 2.

\begin{figure*}[!t]
\centering
\begin{subfigure}{0.65\columnwidth}
\centering\includegraphics[width=\linewidth, ]{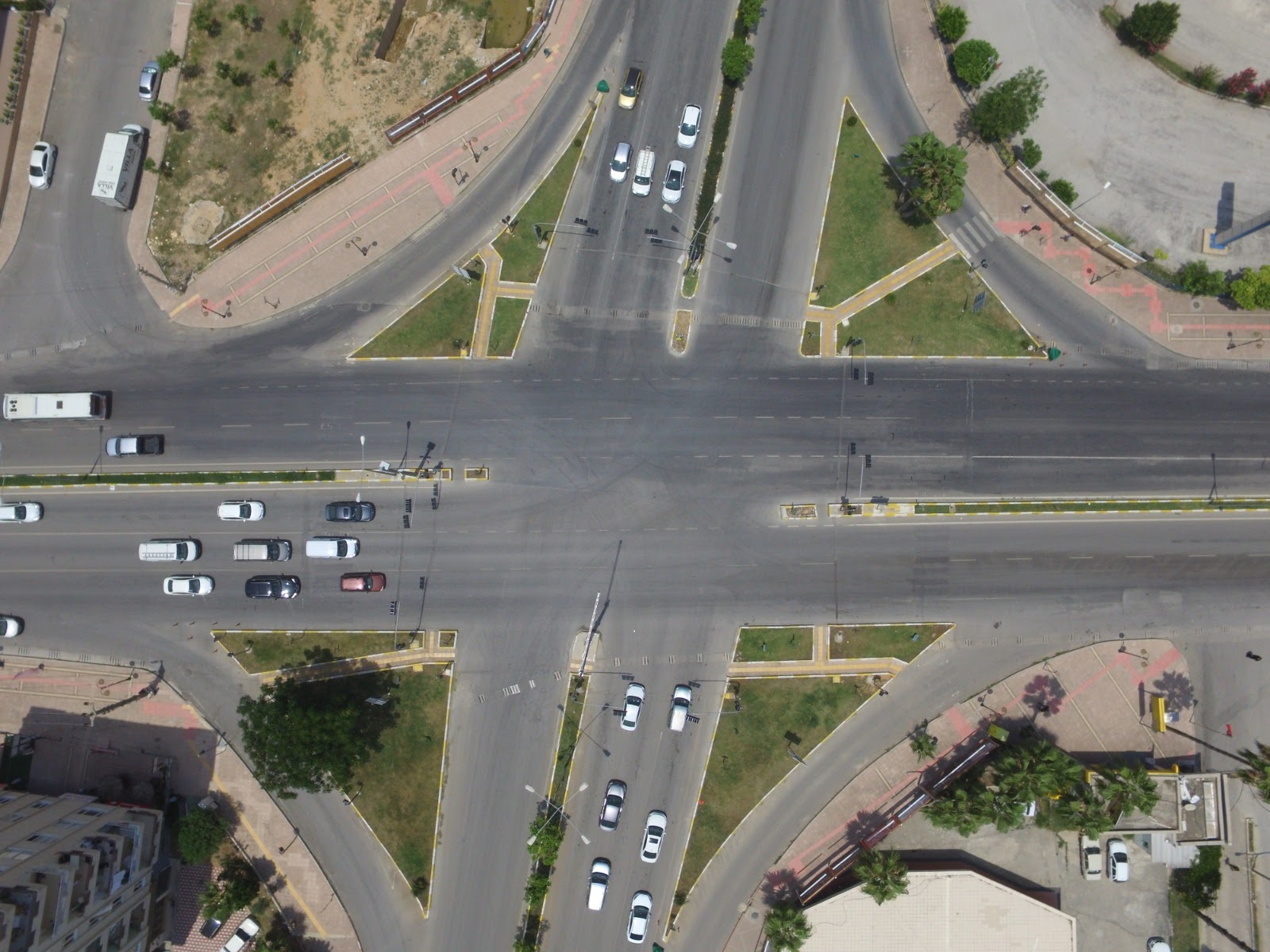}
\caption{}
\end{subfigure}\hfill%%
\begin{subfigure}{0.65\columnwidth}
\centering\includegraphics[width=\linewidth]{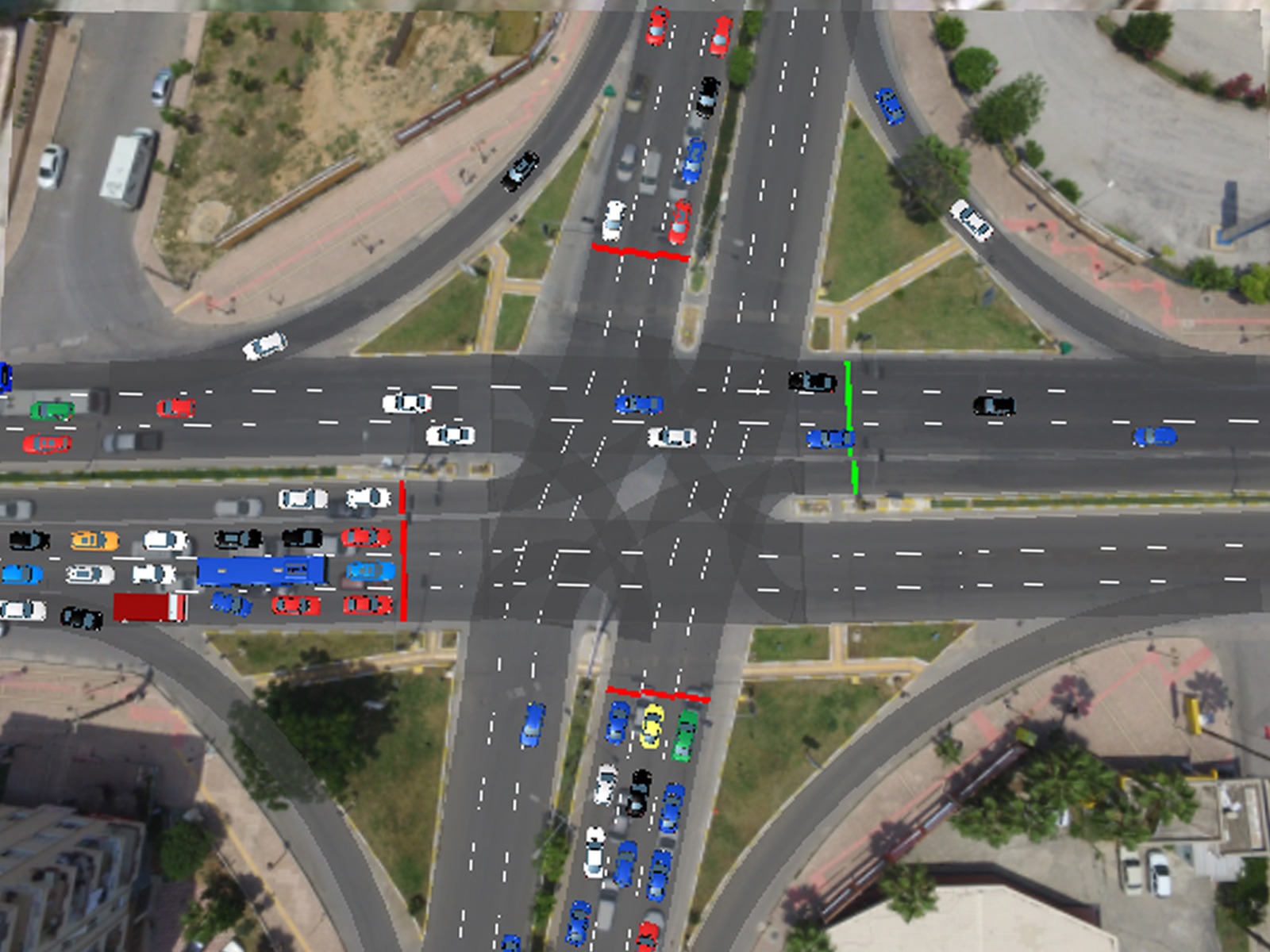}
\caption{}
\end{subfigure}\hfill%
\begin{subfigure}{0.55\columnwidth}
\centering\includegraphics[width=\linewidth]{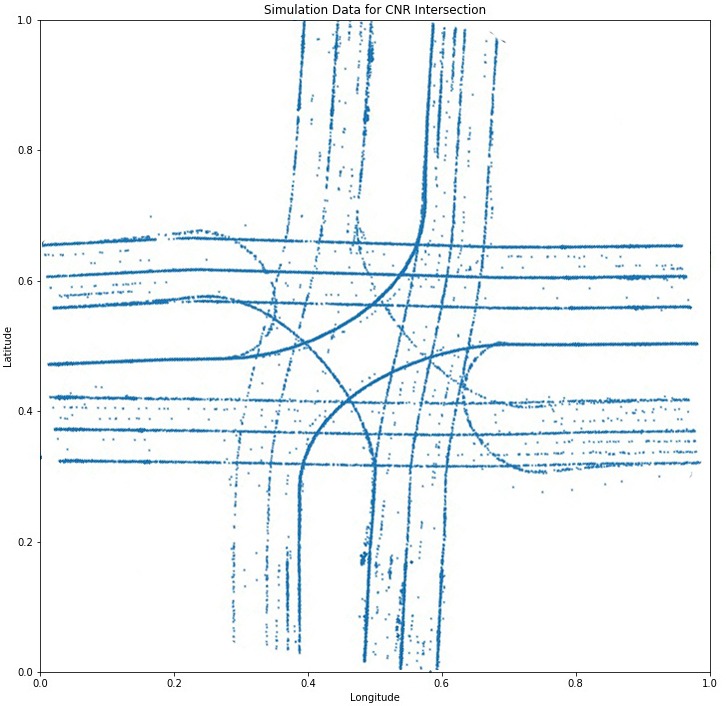}
\caption{}
\end{subfigure}%
\caption{Intersection footage. (a) Drone recorded image of the intersection used in the study. (b) Simulated image of the intersection. (c) Routes in the simulation data.}
\end{figure*}

\subsection{Fisheye Dataset}

We had the access to the installed fisheye surveillance cameras mounted at 8 meters height from the ground on junctions from a city in Turkey. The coverage of the camera reaches a clear human observable view of around 75 meters from the junction entrance point. Each road can accommodate 3 lanes of vehicles that vary in size (car, truck, bus). The obtained dataset \cite{Reference15} is acquired during daytime and sampled at 10fps frames, resulting in 20,000 frames, a sample of the mentioned data is in Figure 3.

\begin{figure}[!h]
\centering
\includegraphics[width=3.2in]{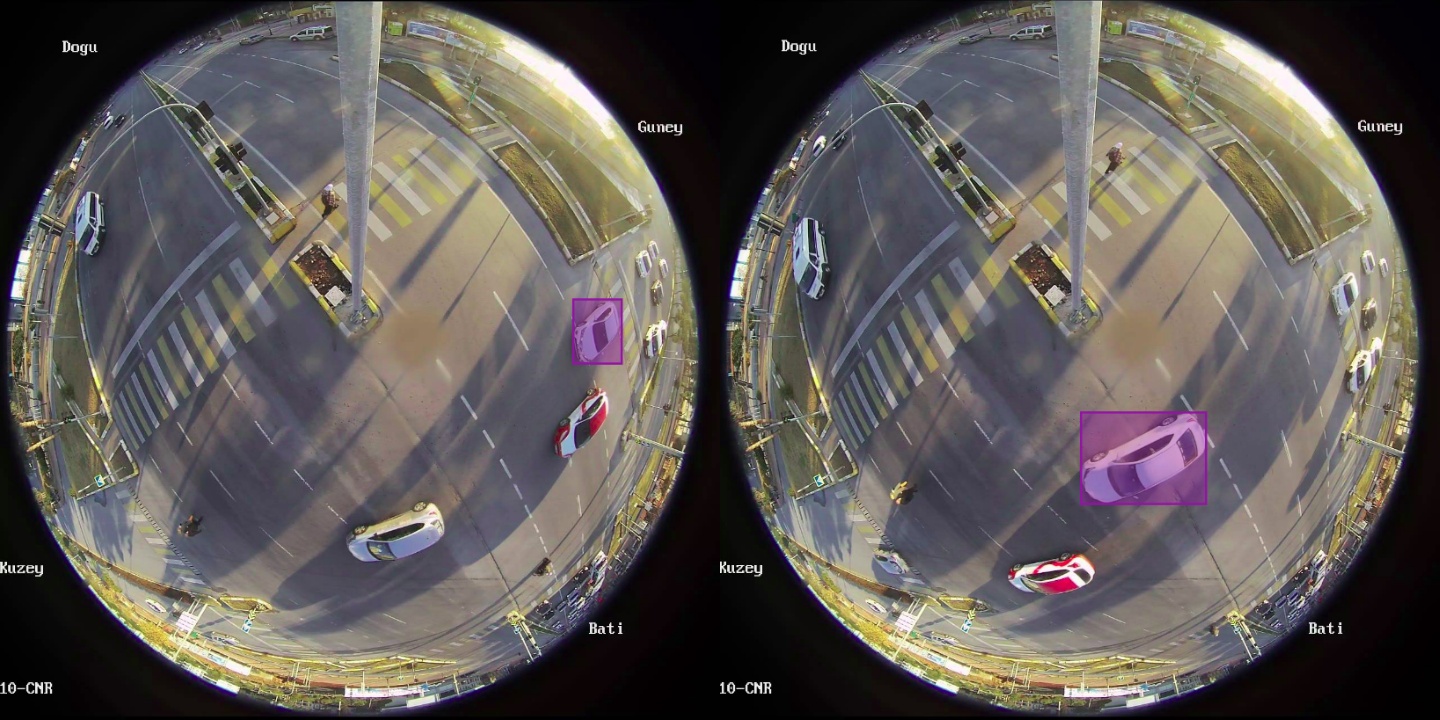}
% where an .eps filename suffix will be assumed under latex, 
% and a .pdf suffix will be assumed for pdflatex; or what has been declared
% via \DeclareGraphicsExtensions.
\caption{Example of the fisheye data.}
\end{figure}

\section{Proposed Method}

\subsection{Vehicle Detection}
Vehicle detection is a variant of the classic object detection problem in computer vision research. Several approaches have addressed detecting vehicles with different study scopes such as autonomous vehicles \cite{R16,R17,R18,R19,R20,R21,R22,R23,R24,R25} and smart cities models \cite{R26,R27,R28,R29,R30,R31,R32,R33,R34}. In our proposed framework, the first step is to detect vehicles at the junction, the capturing is via a static camera with a fisheye lens. 

The fisheye lens has introduced a distortion problem in the vehicle shapes which would fail several techniques used in the research, where standard cameras where used. Some studies tried to overcome the distortion issue with different techniques. In one approach \cite{R35} they have trained a network over the mapping between a fisheye captured lines image and straight lines rectified image to overcome that issue, then let the network auto-rectify any new fisheye incoming image. Affine transformations have also proofed to be a solution to overcome the lens effect \cite{R36,R37}. However, it is considered to be a computationally expensive operation as a pre-process \cite{R38}. Another study \cite{R39} suggests working on the camera calibration itself. Although the found literature is massively informative, the nature of the open environment has limited us from having costly detection pre-processing computation or hardware calibration. 

Here we follow the "use it as it exists" approach to come up with a detection algorithm. We target the fisheye camera as the feed source with no prepossessing taking place, thus eliminating any deployment ambiguities of configuring the algorithm at the post-training stage.

The above-mentioned fisheye data is splitted to train-test subsets 80\% by 20\% respectively. We trained several models on the raw images and tabulated the results in Table-1. The model configuration is cloned from \cite{R40}. It is found that model YoloV3 \cite{R41} overperforms YoloV3-tiny and the Faster-RCNN in the defined scores. Thus, it is selected for the vehicle detection stage. It is well known that relying on deep learning solutions for detection would leverage robustness, but penalize computation time and resources allocation \cite{R42}. 

To evaluate the model, we defined several parameters. For measuring how much punctual the model is we used PASCAL \cite{R43} average precision formula at 3 different Jaccard index (a.k.a IoU) thresholds of interest 50, 75 and 95 \%. The thresholds are picked to get a better understanding of model performance since any vehicle's size in fisheye camera changes massively. For the inference time measurement, the runtime CPU is INTEL core i7-8750H. Due to the existence of several parameters, we defined a formula (Equation-1) to pick up the best model, as illustrated in Table-1.

\begin{align*}
    \mathllap{} score &= 0.4 * mAP_{IoU 50\%} \\
    & \qquad + 0.3 * mAP_{IoU 75\%} \\ 
    &\qquad + 0.1 * mAP_{IoU 95\%} \\
    &\qquad + 0.2 * (1-inference)
\end{align*}

\begin{table}[!h]
\caption{Vehicle detection models}
\begin{adjustbox}{width=\columnwidth}
\begin{tabular}{|c|c|c|c|c|c|}
	\hline
	Model & $mAP_{IoU (50\%)}$ & $mAP_{IoU (75\%)}$ & $mAP_{IoU (95\%)}$  & $Inference_(msec)$ & $score (\%)$\\
	\hline
    YoloV3 & 95.6 & 89.9 & 80.7 & 250 & 88.28\\
    \hline
    YoloV3-tiny  & 89.5 & 80.9 & 55 & 80 & 83.97\\
    \hline
    Faster-RCNN  & 85 & 75 & 70 & 420 & 75.1\\
	\hline
\end{tabular}
\end{adjustbox}
\end{table}

\subsection{Fisheye to Bird's-eye View Conversion}

The wide field of view (FoV) of a fisheye camera benefits in better image coverage comAPred to a standard surveillance camera \cite{R36,R44}. After the vehicles are detected in the fisheye view, it is required to obtain vehicles' trajectories. To limit the computation resources required for this step, it is favored to do a simple pixel transformation mapping on the initial frame and use the obtained transformation parameters to project the trajectories into the bird's-eye view. To do so, as a post-processing step, a geometrical rectification procedure takes place as follows:

\begin{enumerate}
    \item Define $L_f(x_c,y_c)$ as the center point (in pixels) of the vehicle's $V$ bounding box resulted from the detector at fisheye view
    \item Manually dictate the initial frame key points for rectification to bird eye view
    \item Obtain the transformation map matrix
    \item Apply the transformation map matrix against vehicle location $L_f$ to get $L_b$ as the vehicle $V$ location at the bird's-eye view; we denote $L_b$ as our $L$ for the sake of simplicity.
\end{enumerate}

The algorithm of distortion model estimation is described  in Equation-2:

\begin{equation}
D(P_f,P_c)=K(z)*D(P_b,P_c).
\end{equation}

where \emph{D(a,b)} is defined as $\begin{pmatrix}a_x-b_x\\a_y-b_y\end{pmatrix}$. $P_c$ is defined to be the center point of distortion which is the center of the image frame, $P_f$ is the point $P$ in the fisheye view while $P_b$ is the point $P$ in the bird's-eye view. In order to obtain the trajectory point $P_b$ of $P_f$ it is required to solve the Equation-3 for $K(z)$ where $K(z)$ is the function approximating the distortion as a Taylor series defined as:

\begin{equation} 
K(z) = q_0 + q_1z + q_2z^2 + q_3z^3 + \cdots 
\end{equation}

The $z$ is the Euclidean distance between $P_c$ and $P_f$:

\begin{equation}
z=\sqrt{(P_{cx} - P_{fx})^2 + (P_{cy} - P_{fy})^2}.
\end{equation}

Solving for $K$ requires obtaining the distortion parameters set $(q_0,q_1,q_2,...)$. This is done at the initially obtained frame, where a line of points $P_f$ is drawn on the fisheye image following the path of the street, as in Figure 4. Lastly using the obtained distortion parameters we are able to project each vehicle location $L_f$ to $L_b$.

\begin{figure}[!h]
\centering
\includegraphics[width=3.2in]{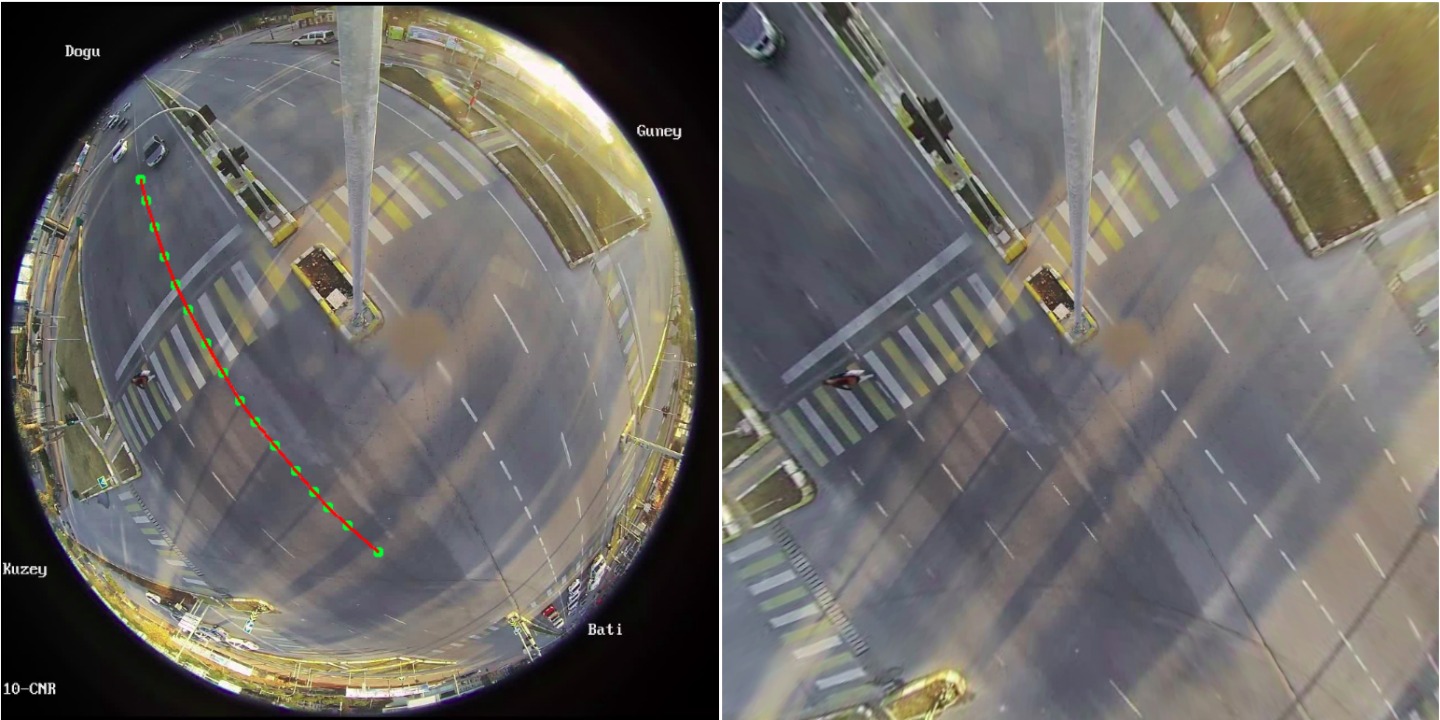}
\caption{On the left is a fisheye frame having a drawn line of points $P_f$ is defined to follow the street path, on the right is the rectified bird's-eye version of the frame}
\end{figure}

\subsection{Tracking the vehicle trajectory}
To define the vehicle trajectory it is required to track each passing vehicle throughout the junction. Research community has addressed object tracking problem with geometrical algebra \cite{R45,R46,R47,R48,R49} as well as recent deep learning-based solution \cite{R50,R51,R52}. The requirement to have a lightweight solution for trajectory extraction emphasizes going with the Munkres \cite{R53} Kalman \cite{R54} combination, thus achieving high accuracy and computationally efficient solution. 

Several studies have addressed object association using the Munkres algorithm. In the study of body parts association of COCO 2016 challenge Cao et al. have suggested the Munkres algorithm for optimal results \cite{R46}. Others like Mozaffari et al. have used it in Unmanned Autonomous Vehicle (UAV) deployment and association problems in order to keep track of the devices \cite{R47}.

The Munkres is to link identified objects at frame $F_{t-1}$ and frame $F_t$. To utilize it in our vehicle association and have a track of the vehicles' Spatio-temporal status, we have favored the conditional context to be the Jaccard Index score. Although some addressed studies have preferred to use Euclidean distance \cite{R48} the fact of vehicles can get too close may trick the algorithm at that point.

\begin{equation}
    J_{t-1,t} = \begin{pmatrix}
    j_{1_{t-1},1_t}&j_{1_{t-1},2_t}&j_{1_{t-1},3_t}
    \\
    j_{2_{t-1},1_t}&j_{2_{t-1},2_t}&j_{2_{t-1},3_t}
    \\
    j_{3_{t-1},1_t}&j_{3_{t-1},2_t}&j_{3_{t-1},3_t}
    \end{pmatrix}
\end{equation}
where $j_{t-1,t} = Jaccard(V_{t-1},V_t)$ and $V$ is the vehicle detected at location $L$ in the previous step. The method will associate the vehicles by obtaining the maximum IoU $Max(J_{t-1,t})$, as illustrated in Figure 5.

\begin{figure}[!h]
\centering
\includegraphics[width=3.2in]{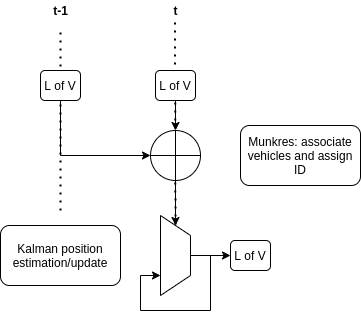}
% where an .eps filename suffix will be assumed under latex, 
% and a .pdf suffix will be assumed for pdflatex; or what has been declared
% via \DeclareGraphicsExtensions.
\caption{Vehicle $V$ trajectory tracking procedure}
\end{figure}

After assigning the vehicles, it is required to track their path throughout the junction. The more the path tracking mechanism is consistent, the better the proposed method performance. Kalman filter has proved to be a state of the art tracking method, deployed in several studies \cite{R48,R55,R56,R57}. We use Kalman filter to estimate vehicle $V$ next location $L_t$ based on $L_{t-1}$ and smooth the trajectory produced in case the detector failed to detect $V$ at a frame $F$. The vehicle location $L$ is defined for the tracker when the $V$ enters the camera FoV and is dismissed when it leaves the FoV. The output of this stage can be formulated as below:
\begin{equation}
    \hat{V}=Kalman(Munkres(V_{@L}))
\end{equation}
where $V\in\{V_{0@L0}, V_{1@L1},...\}$ and $\hat{V}$ denotes the set of tracked vehicles in the scene at time $t$.

\subsection{Route Classification}

At the initial stage, each route in the junction is drawn manually over the image frame. The route line consists of 8 to 10 connected points are fitted to 2 to 7-degree polynomials as shown in Figure 6-a and 6-b.  

\begin{figure}[!h]
\centering
\subfloat[]{\label{a1}\includegraphics[width=.48\linewidth]{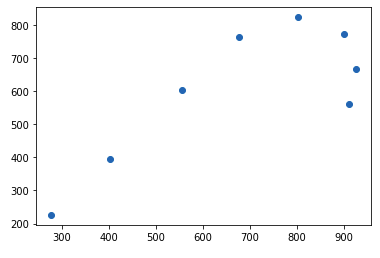}}\hfill
\subfloat[]{\label{b1}\includegraphics[width=.48\linewidth]{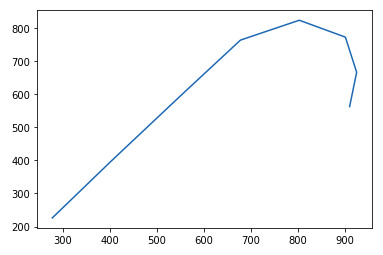}}\par 
\caption{An example route. (a) The points for the route. (b) The related line of the route.}
\label{fig2}
\end{figure}

After that, the mean absolute errors are calculated to correlate the vehicle $V$'s trajectory with the defined routes, with the formula in Equation-7. Finally, the vehicle $V$ belongs to the route with the lowest error.

\begin{equation} 
error = \frac{\sum_{i}^{n}\lvert P_{by_i} - \hat{y_i} \rvert}{n}.
\end{equation}

This process is done for each route. An example route classification process is shown below in Figure 7. The original and bird-eye view of fisheye frame and classified routes are shown in Figure 8.

\begin{figure}[!h]
\centering
\includegraphics[width=3in,keepaspectratio]{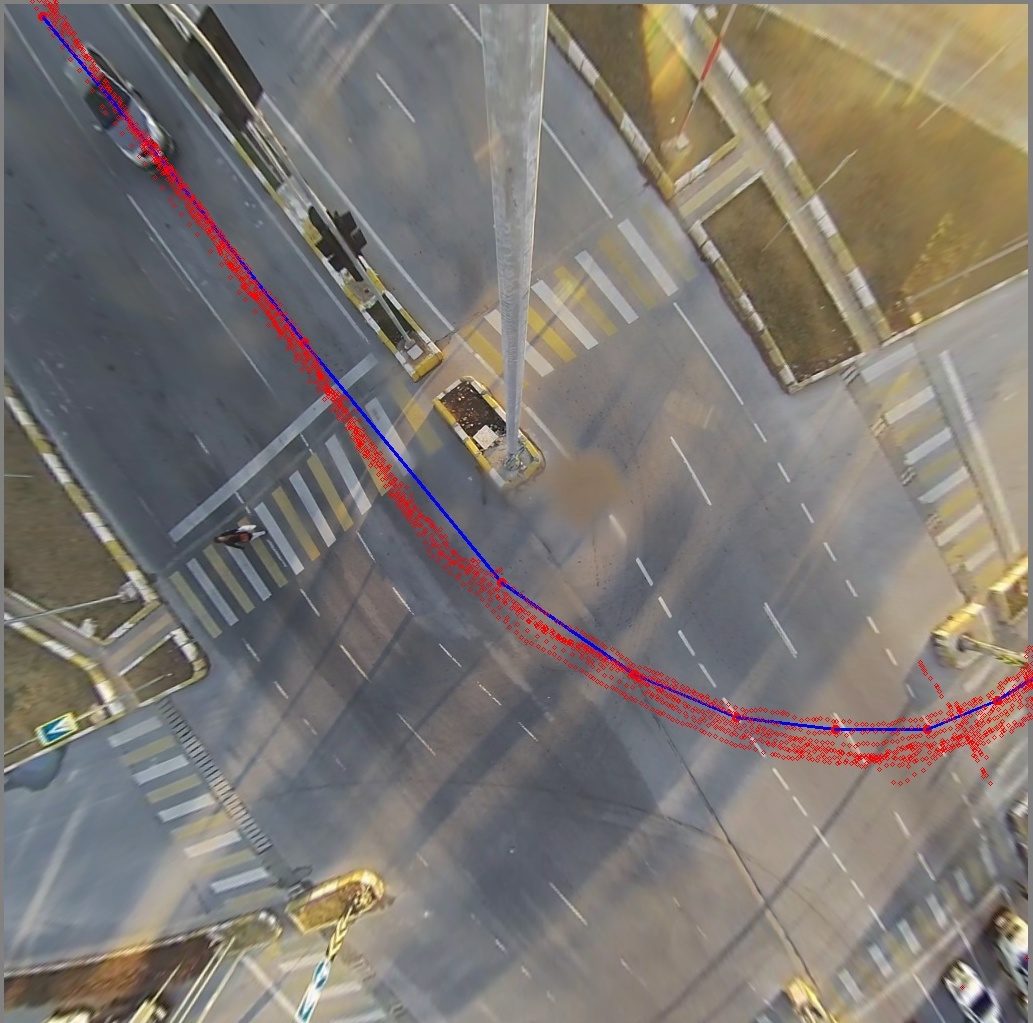}
% where an .eps filename suffix will be assumed under latex, 
% and a .pdf suffix will be assumed for pdflatex; or what has been declared
% via \DeclareGraphicsExtensions.
\caption{A sample of route classification for one route. Red ones are the followed trajectory and blue line is the manually dictated route}
\end{figure}

\begin{figure}[!h]
\centering
\begin{subfigure}{4cm}
\centering\includegraphics[width=3.5cm]{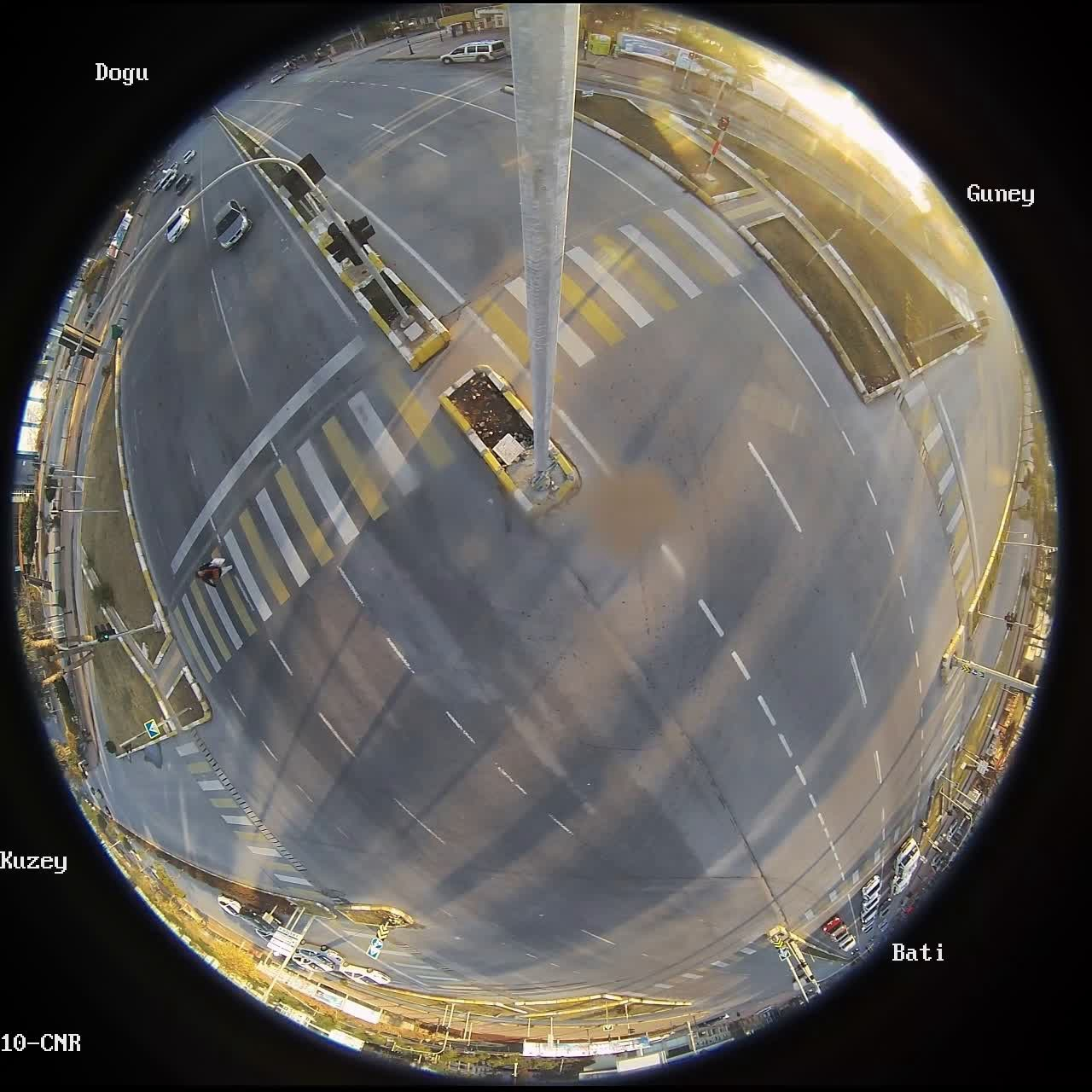}
\caption{}
\end{subfigure}%
\begin{subfigure}{4cm}
\centering\includegraphics[width=3.5cm]{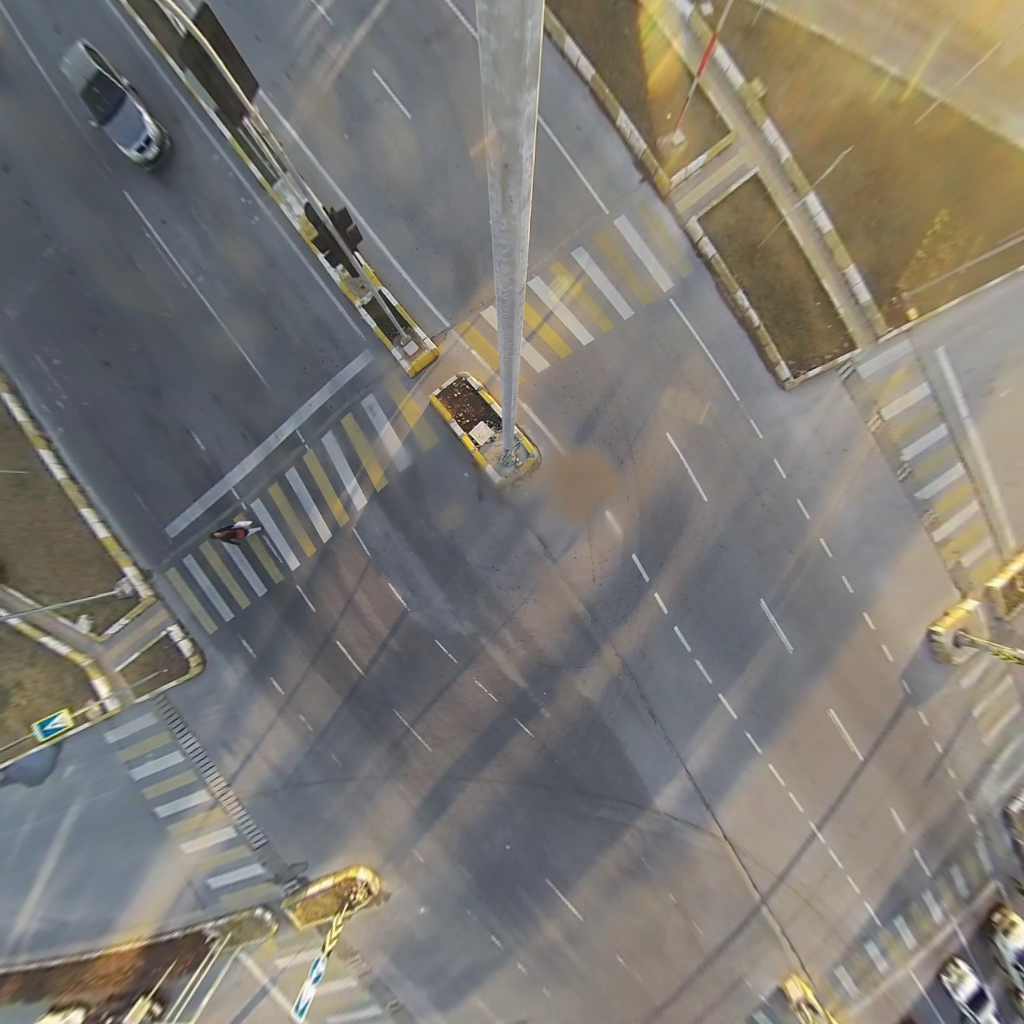}
\caption{}
\end{subfigure}\vspace{10pt}
 
\begin{subfigure}{4cm}
\centering\includegraphics[width=3.5cm]{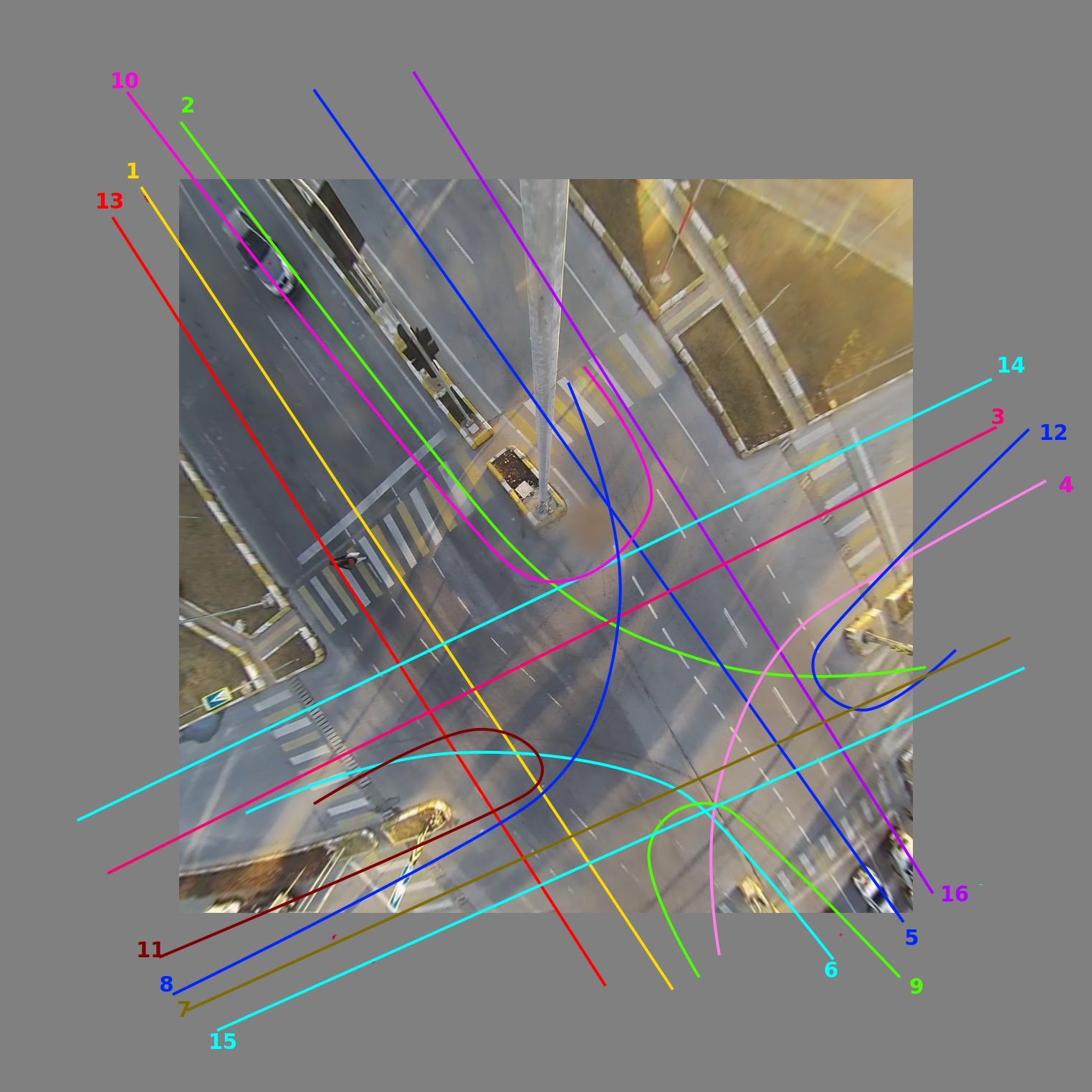}
\caption{}
\end{subfigure}%
\begin{subfigure}{4cm}
\centering\includegraphics[width=3.5cm]{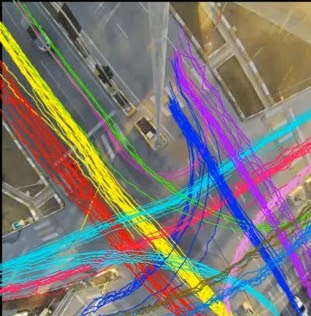}
\caption{}
\end{subfigure}
\caption{Example of a fisheye frame and bird eye view. (a) The fisheye frame. (b) The bird-eye view. (c) Route directions. (d) Classified Routes.}
\end{figure}

\subsection{Anomaly Detection}

\subsubsection{Setup}
The vehicle coordinates coming from the fisheye camera are sorted according to time, and the route followed by the vehicle is determined. The routes of all vehicles passing in a day are collected. These routes are divided into classes for each direction of the junction. Classified routes of vehicles at the end of the day are modeled by polynomial functions with different degrees from 1 to 20 chosen by the least-squares method. 

The error value of each modeling for each vehicle is recorded, and then the average error values of these models for each degree are calculated. These average error values are comAPred with the defined error threshold. The lowest degree having the average error value below the error threshold is recorded for each route class.

\begin{figure*}[!h]
\centering
\begin{subfigure}{\columnwidth}
\centering\includegraphics[width=\linewidth]{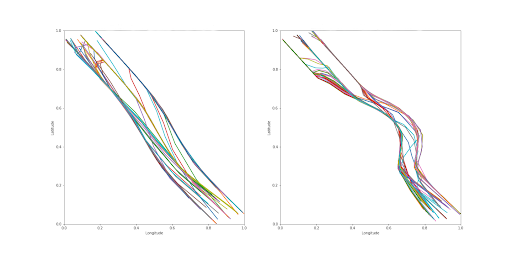}
\caption{}
\end{subfigure}%
\begin{subfigure}{\columnwidth}
\centering\includegraphics[width=\linewidth]{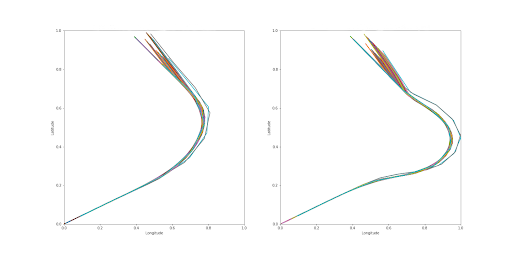}
\caption{}
\end{subfigure}\vspace{10pt}
\caption{Examples of a deviation in the route for simulation data. Figures (a) and (b) represent different routes.}
\end{figure*}

\begin{figure*}[!h]
\centering
\includegraphics[width=6.4in]{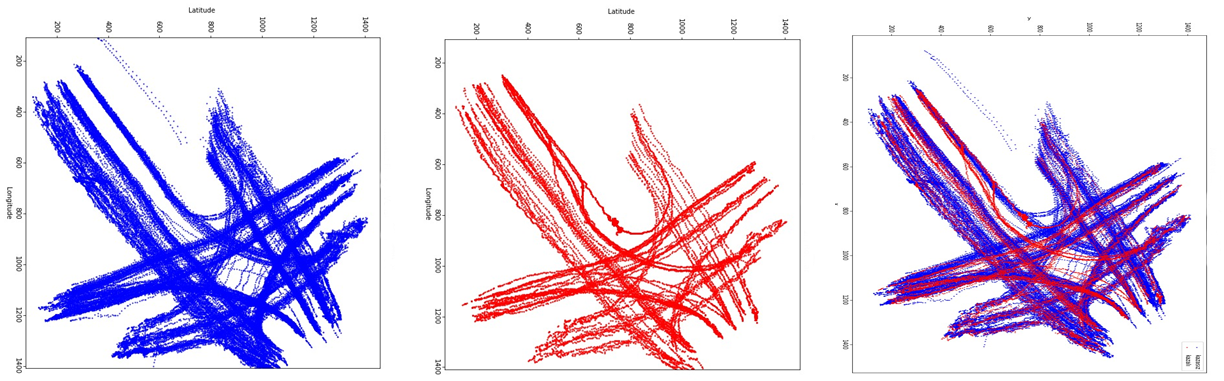}
% where an .eps filename suffix will be assumed under latex, 
% and a .pdf suffix will be assumed for pdflatex; or what has been declared
% via \DeclareGraphicsExtensions.
\caption{Example of a deviation in the route for fisheye data.}
\end{figure*}

\subsubsection{Runtime}

The next day, after determining the class of the incoming data by using the minimum mean square error, it is merged with the data of the past 5 vehicles in the same class. Merged data is modeled by the polynomial functions with different degrees from 1 to 20, the average error value is calculated, and the lowest degree is comAPred with the previous day's recorded lowest degree to determine whether there is a deviation in the route. Examples of the deviations in the route for simulation data and fisheye data are shown in Figures 9 and 10.

If the degree difference between the lowest degrees is more than 2, it is classified as an anomaly. This threshold value is decided by the box-plot extreme value detection procedure. All quartiles of the simulation data are calculated. Then, the interquartile range (IQR) and whisker lengths are obtained as follows: The upper whisker of the box plot is the largest data point smaller than 1.5*IQR above the third quartile, and the lower whisker of the box plot is the smallest data point larger than 1.5*IQR below the first quartile. If the newly observed data is below the lower whisker or above the upper whisker, it is labeled as an anomaly. In our simulation data, the threshold value to decide anomaly is calculated as the 2-degree difference between the fitted polynomial functions. The IQR method does not depend on any loops and is, therefore, faster and more easily scaled \cite{R58}. An example of how this anomaly detection works can be seen in Figure 11. Here, the second vehicle on the route made an accident, and this method catches it immediately.

\begin{figure}[!h]
\centering
\includegraphics[height=5cm, keepaspectratio]{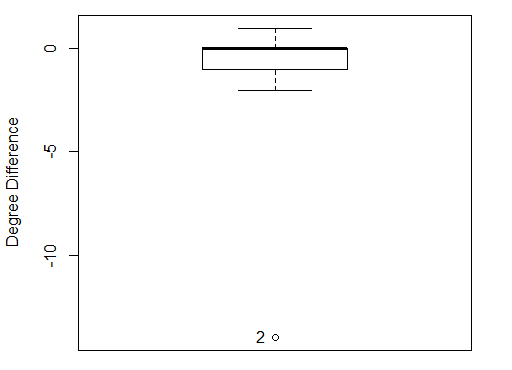}
\caption{The box-plot of the degree differences between the simulated and real data having an accident.}
\end{figure}

To support the decision of degree threshold, a simulation scenario that has no incident was run for 30 times to characterize the normal behavior of the trajectories. Several resources \cite{R59,R60} were used to decide how many times the simulation should be run. After simulating for 30 times, two route types were selected to analyze the degree change between simulations. Selected routes and related results are shown below in Figure 12 and Table 2. As it is seen from the results, the chosen difference threshold value is supported by the simulation results.

\begin{figure}[!h]
\centering
\subfloat[]{\label{a}\includegraphics[width=.48\linewidth]{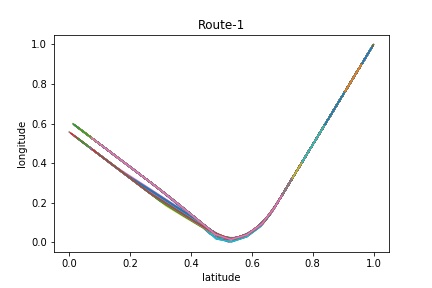}}\hfill
\subfloat[]{\label{b}\includegraphics[width=.48\linewidth]{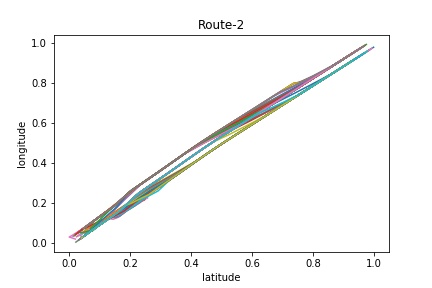}}\par 
\caption{Selected Routes for Analyzing Simulation Data. (a) Turning route (b) Straight route}
\label{fig3}
\end{figure}

\begin{table}[!h]
\caption{Lowest degree results for simulation data}
\begin{adjustbox}{width=\columnwidth}
\begin{tabular}{|c|c|c|c|}
	\hline
	Route & 10 percentile of Lowest Degree & Median & 90 percentile of Lowest Degree\\
	\hline
    Route-1 & 3 & 4 & 4\\
	\hline
    Route-2 & 1 & 1 & 1\\
	\hline
\end{tabular}
\end{adjustbox}
\end{table}

\section{Conclusion}

In this study, a framework for incident detection at an intersection is developed. An omnidirectional (fisheye) camera acts as a data source, and based on the vehicles' trajectories, anomaly occurrence decision is made.  The evaluation measures for algorithms are shown below in Table 3. Since the proposed method includes merging the incoming data with the data of the past 5 vehicles, there are wrongly classified vehicle trajectories because of taking the average residue values for these 6 vehicles. However, we found that all routes with deviations are successfully detected.

\begin{table}[!h]
\caption{Anomaly detection evaluation measures}
\begin{adjustbox}{width=\columnwidth}
\begin{tabular}{|c|c|c|c|c|}
	\hline
	Algorithm & Accuracy(\%) & Precision(\%) & Recall(\%) & F1-Score(\%)\\
	\hline
    Anomaly Detection (Synthetic data) &  96.8 & 100 & 80 & 89 \\
	\hline
	Anomaly Detection (Fisheye data - Vehicle Based) & 97.3 & 64 & 39 & 48.5\\
	\hline
	Anomaly Detection (Fisheye data - Route Based) & 90.8 & 33.3 & 100 & 50\\
	\hline
\end{tabular}
\end{adjustbox}
\end{table}

The proposed method does not include u-turns. Therefore, these turns are not taken into account for the evaluation. It is planned to carry out a future study including the u-turns. It is also planned to identify and analyze the roads joining the intersection where the incident takes place.

% biography section
% 
% If you have an EPS/PDF photo (graphicx package needed) extra braces are
% needed around the contents of the optional argument to biography to prevent
% the LaTeX parser from getting confused when it sees the complicated
% \includegraphics command within an optional argument. (You could create
% your own custom macro containing the \includegraphics command to make things
% simpler here.)
%\begin{IEEEbiography}[{\includegraphics[width=1in,height=1.25in,clip,keepaspectratio]{mshell}}]{Michael Shell}
% or if you just want to reserve a space for a photo:

\begin{IEEEbiographynophoto}{Murat Tulgaç} received the B.S. degree in electrical and electronics engineering from Anadolu University, Eskişehir, Turkey, in 2018. He is currently pursuing the M.S. degree in electrical and electronics engineering with TOBB ETÜ, Ankara, Turkey. He is also working as Data Scientist at ISSD. His research interests include machine learning, time series analysis, and distributed computing.
\end{IEEEbiographynophoto}

% if you will not have a photo at all:
\begin{IEEEbiographynophoto}{Enes Yüncü} received the B.S. degree in electrical and electronics engineering and the M.S. degree in cognitive science from the METU, Ankara, Turkey, in 2010 and 2013, respectively. He is currently working as Technical Lead at ISSD. His research interests include deep learning, image processing, and computer vision.
\end{IEEEbiographynophoto}

\begin{IEEEbiographynophoto}{Mohammad Alhaddad} received a B.Sc degree in mechatronics engineering from Eastern Mediterranean University in 2017. He is with the ISSD image processing team since 2018 with a focus on machine intelligence, software development, and design aspects of machine learning-based projects. He has over 3 years of research experience in the sustainability aspects of IoT systems and has worked on a Tunnel surveillance deep learning empowerment project funded by the EU in 2019-2020.
\end{IEEEbiographynophoto}

% insert where needed to balance the two columns on the last page with
% biographies

\begin{IEEEbiographynophoto}{Ceylan Yozgatlıgil} received the M.Sc. degree from the Department of Statistics, METU, in 1999, and the Ph.D. degree in Statistics from the Temple University in 2007. She is currently an Associate Professor in the Department of Statistics at METU. Her research interests include time series analysis, temporal aggregation of time series, applied statistics, supervised learning via machine learning algorithms, statistical applications in meteorological variables, temporal clustering methods, statistical inference.
\end{IEEEbiographynophoto}

% You can push biographies down or up by placing
% a \vfill before or after them. The appropriate
% use of \vfill depends on what kind of text is
% on the last page and whether or not the columns
% are being equalized.

%\vfill

% Can be used to pull up biographies so that the bottom of the last one
% is flush with the other column.
%\enlargethispage{-5in}

% that's all folks
\end{document}